%% file: IEEEdest08di.tex
\documentclass[conference,a4paper]{IEEEtran}
\usepackage{graphicx, acronym, color, url}
\usepackage{ifthen}
\usepackage{substr}
\usepackage{lastpage, graphicx, color, float, setspace}

\usepackage[
pdftitle={Digital Ecosystems: Self-Organisation of Evolving Agent Populations}, 
pdfauthor={Gerard Briscoe},
pdfsubject={Ecosystem-Oriented Architectures},
colorlinks=true, linkcolor=black, citecolor=black,  filecolor=black, urlcolor=black,
hyperfootnotes=false]{hyperref} 

\newboolean{final}\setboolean{final}{true}

\newcommand{\tfigure}[9]
	{
	\IfSubStringInString{!}{#7}{\begin{figure}[#7]}{\begin{figure}[!t]}
	\IfSubStringInString{mm}{#8}{\vspace{#8}}{}
	\centering
	
	\IfSubStringInString{pdf}{#3}
		{
		\execute{cd images; ln -s #2.pdf .#2.gdf}
		\href{file://localhost/Users/g/Library/Application\ Support/Interarchy/Net\ Disks/gSpace/PastPapers/IEEEdest2008-di/images.#2.gdf}{\includegraphics[#1]{images/#2}}
		}
		{\IfSubStringInString{graph}{#3}
			{
			\execute{cd images; ./makeGraph.sh #2; ln -s #2.pdf .#2.gdf}
			\ifthenelse{\boolean{final}}
				{\includegraphics[#1]{images/#2}}
				{\href{file://localhost/Users/g/Library/Application\ Support/Interarchy/Net\ Disks/gSpace/PastPapers/IEEEdest2008-di/images/.#2.gdf}{\includegraphics[#1]{images/#2}}}
			}
			{
			\execute{cd images; ./pdfcrop.sh #2}
			\ifthenelse{\boolean{final}}
				{\includegraphics[#1]{images/#2-crop.pdf}}
				{\href{file://localhost/Users/g/Library/Application\ Support/Interarchy/Net\ Disks/gSpace/PastPapers/IEEEdest2008-di/images/#2.#3}{\includegraphics[#1]{images/#2-crop.pdf}}}
			}
		}
		
	\vspace{#6}
	\caption[#4]
		{
		\label{#2}
		\tcaption{#4}{#5}
		}
	\IfSubStringInString{mm}{#9}{\vspace{#9}}{}
	\end{figure}
	}

\newcommand{\mfigure}[7]
	{
	\vspace{#5}
		\IfSubStringInString{!}{#7}
			{\begin{figure}[#7]}{\begin{figure}[!t]}
		#1
		\vspace{#6}
		\caption[#2]{\italics{#2: #3}}
		\label{#4} 
	\end{figure}
	}

\definecolor{tred}{RGB}{255,0,0} 
\definecolor{tblue}{RGB}{102,204,255} 
\definecolor{red}{RGB}{128,0,0} 
\definecolor{blue}{RGB}{0,0,128} 
\definecolor{green}{RGB}{0,128,0} 
\definecolor{yellow}{RGB}{128,128,0} 
\definecolor{purple}{RGB}{128,0,128} 
\definecolor{turquoise}{RGB}{0,128,128} 
\definecolor{grey}{RGB}{76,76,76}
\definecolor{brown}{RGB}{128,64,0}

\ifthenelse{\boolean{final}}
	{}
	{}
\newcommand{\red}[1]{\color{red}#1\normalcolor}
\newcommand{\blue}[1]{\color{blue}#1\normalcolor}
\newcommand{\green}[1]{\color{green}#1\normalcolor}
\newcommand{\yellow}[1]{\color{yellow}#1\normalcolor}
\newcommand{\purple}[1]{\color{purple}#1\normalcolor}
\newcommand{\turquoise}[1]{\color{turquoise}#1\normalcolor}
\newcommand{\grey}[1]{\color{grey}#1\normalcolor}
\newcommand{\brown}[1]{\color{brown}#1\normalcolor}

\newcommand{\italics}{\textit}
\newcommand{\bold}{\textbf}

\newcommand{\tcaption}[2]
	{
	\IfSubStringInString{:}{#2}{\italics{#1 #2}}{\italics{#1: #2}}
	}

\newcommand{\execute}[1]{\immediate\write18{#1}}
\newcommand{\setCap}[2]{#1\immediate\write18{./mkcaption.sh #2}}
\newcommand{\getCap}[1]{\acl{#1}}

\begin{document}
\input{acronyms}

\input{captions}
\immediate\write18{echo > captions.tex}

\title{Digital Ecosystems: \\Optimisation by a Distributed Intelligence}

\author{
\authorblockN{Gerard Briscoe}
\authorblockA{Digital Ecosystems Lab\\
Department of Media and Communications\\
London School of Economics and Political Science\\
London, United Kingdom\\
e-mail: g.briscoe@lse.ac.uk}
\and
\authorblockN{Philippe De Wilde}
\authorblockA{Intelligent Systems Laboratory\\
School of Mathematical and Computer Sciences\\
Heriot-Watt University\\
Edinburgh, United Kingdom\\
e-mail: pdw@macs.hw.ac.uk}
}
\maketitle

\begin{abstract}

Can intelligence optimise Digital Ecosystems? How could a distributed intelligence interact with the ecosystem dynamics? Can the software components that are part of genetic selection be intelligent in themselves, as in an adaptive technology? We consider the effect of a \emph{distributed intelligence} mechanism on the evolutionary and ecological dynamics of our Digital Ecosystem, which is the \emph{digital} counterpart of a biological ecosystem for evolving software services in a distributed network. We investigate \acp{NN} and \ac{SVM} for the learning based pattern recognition functionality of our \emph{distributed intelligence}. Simulation results imply that the Digital Ecosystem performs better with the application of a \emph{distributed intelligence}, marginally more effectively when powered by \ac{SVM} than \acp{NN}, and suggest that it can contribute to optimising the operation of our Digital Ecosystem.\\


\end{abstract}
\IEEEpeerreviewmaketitle

\section{Introduction}

With a Digital Ecosystem, being the \emph{digital} counterpart of a biological ecosystem for evolving software services in a distributed network, can we answer the following questions; Can intelligence optimise the evolutionary process? How could a distributed intelligence interact with the ecosystem dynamics? Can the software components that are part of genetic selection be intelligent in themselves, as in an adaptive technology? These are wide ranging questions, and we have started by considering a \emph{distributed intelligence} based on a simple social interaction mechanism that leads to \emph{targeted migration}. We will use a \emph{machine learning} technique to power our \emph{distributed intelligence}, considering both \acp{NN} and \ac{SVM}. We will start with a brief reminder for our definition of Digital Ecosystems.

\label{simRes}

Our Digital Ecosystem \cite{javaOne} is the digital counterpart of a natural ecosystem \cite{dbebook, de07oz}, which automates the search for new algorithms in a scalable architecture, through the evolution of software services in a distributed network. In the Digital Ecosystem, local and global optimisations concurrently operate to determine solutions to satisfy different optimisation problems. The global optimisation here is not a decentralised super-peer based control mechanism \cite{risson2006srt}, but the completely distributed peer-to-peer network of the interconnected habitats, which are therefore not susceptible to the failure of super-peers. This is a novel optimisation technique inspired by biological ecosystems, working at two levels: a first optimisation, migration of agents which are distributed in a peer-to-peer network, operating continuously in time; this process feeds a second optimisation, based on evolutionary combinatorial optimisation, operating locally on single peers and is aimed at finding solutions that satisfy locally relevant constraints. So, the local search is improved through this twofold process to yield better local optima faster, as the distributed optimisation provides prior sampling of the search space through computations already performed in other peers with similar constraints \cite{bionetics}. The services consist of an executable component and a descriptive semantic component. Analogous to the way in which an agent is capable of execution and has an ontological description. So, if the services are modelled as software agents \cite{D6.1}, then the Digital Ecosystem can be considered a \ac{MAS} which uses distributed evolutionary computing to combine suitable agents available to meet user requests for applications. 

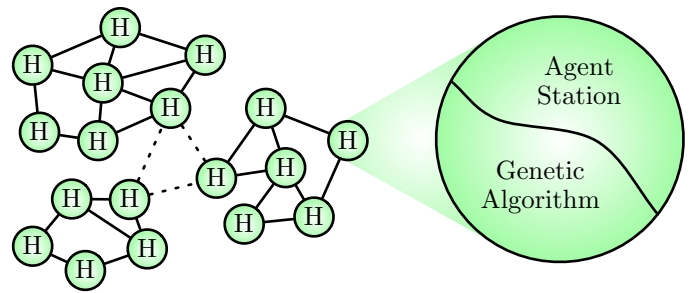
\tfigure{width=3.5in}{architecture2}{graffle}{Digital Ecosystem}{Optimisation architecture in which agents travel along the peer-to-peer connections; in every node (habitat) local optimisation is performed through an evolutionary algorithm, where the search space is determined by the agents present at the node.}{-7mm}{}{}{}

The motivation for using parallel or distributed evolutionary algorithms is twofold: first, improving the speed of evolutionary processes by conducting concurrent evaluations of individuals in a population; second, improving the problem-solving process by overcoming difficulties that face traditional evolutionary algorithms, such as maintaining diversity to avoid premature convergence \cite{muhlenbein1991eta, stender1993pga}. The fact that evolutionary computing manipulates a population of independent solutions actually makes it well suited for parallel and distributed computation architectures \cite{cantupaz1998spg}. There are several variants of distributed evolutionary computing, leading some to propose a taxonomy for their classification \cite{nowostawski1999pga}, with there being two main forms \cite{cantupaz1998spg, stender1993pga}: multiple-population/coarse-grained migration/island models \cite{lin1994cgp, cantupaz1998spg}, and single-population/fine-grained diffusion/neighbourhood models \cite{manderick1989fgp, stender1993pga}. Fine-grained \emph{diffusion} models \cite{manderick1989fgp, stender1993pga} assign one individual per processor. A local neighbourhood topology is assumed, and individuals are allowed to mate only within their neighbourhood, called a \emph{deme}\footnote{In biology a deme is a term for a local population of organisms of one species that actively interbreed with one another and share a distinct gene pool \cite{devisser2007ese}.}. The demes overlap by an amount that depends on their shape and size, and in this way create an implicit migration mechanism. Each processor runs an identical evolutionary algorithm which selects parents from the local neighbourhood, produces an offspring, and decides whether to replace the current individual with an offspring. In the coarse-grained \emph{island} models \cite{lin1994cgp, cantupaz1998spg}, evolution occurs in multiple parallel sub-populations (islands), each running a local evolutionary algorithm, evolving independently with occasional \emph{migrations} of highly fit individuals among sub-populations. This model has also been used successfully in the determination of investment strategies in the commercial sector, in a product known as the Galapagos toolkit \cite{galapagos1, galapagos2}. However, all the \emph{islands} in this approach work on exactly the same problem, which makes it less analogous to biological ecosystems in which different locations can be environmentally different \cite{begon96}. 

The landscape, in energy-centric biological ecosystems, defines the connectivity between habitats \cite{begon96}. Connectivity of nodes in the digital world is generally not defined by geography or spatial proximity, but by information or semantic proximity. For example, connectivity in a peer-to-peer network is based primarily on bandwidth and information content, and not geography. The island-models of \acl{DEC} use an information-centric model for the connectivity of nodes (\emph{islands}) \cite{lin1994cgp}. However, because it is generally defined for one-time use (to evolve a solution to one problem and then stop) it usually has a fixed connectivity between the nodes, and therefore a fixed topology \cite{cantupaz1998spg}. So, supporting evolution in the Digital Ecosystem, with a dynamic multi-objective \emph{selection pressure} (fitness landscape \cite{wright1932} with many peaks), requires a re-configurable network topology, such that habitat connectivity can be dynamically adapted based on the observed migration paths of the agents between the users within the habitat network. So, based on the island-models of \acl{DEC} \cite{lin1994cgp}, each connection between the habitats is bi-directional and there is a probability associated with moving in either direction across the connection, with the connection probabilities affecting the rate of migration of the agents. However, additionally, the connection probabilities will be updated by the success or failure of agent migration using the concept of Hebbian learning \cite{hebb}: the habitats which do not successfully exchange agents will become less strongly connected, and the habitats which do successfully exchange agents will achieve stronger connections. This leads to a topology that adapts over time, resulting in a network that supports and resembles the connectivity of the user base, typically small-world networks \cite{white2002nst, antionella}. They have \setCap{many strongly connected clusters (communities), called \emph{sub-networks} (quasi-complete graphs), with a few connections between these clusters (communities) \cite{swn1}. Graphs with this topology have a very high clustering coefficient and small characteristic path lengths \cite{swn1},}{archComTop} as visualised in Figure \ref{architecture2}. The novelty comes from the creation of multiple evolving populations in response to \emph{similar} requests, whereas in the island-models of \acl{DEC} there are multiple evolving populations in response to only one request \cite{lin1994cgp}. So, in our Digital Ecosystem different \setCap{requests are evaluated on separate \emph{islands} (populations), with their evolution accelerated by the sharing of solutions between the evolving populations (islands), because they are working to solve similar requests (problems).}{similarCap}

The users \setCap{will formulate queries to the Digital Ecosystem by creating a request as a \emph{semantic description}, like those being used and developed in \aclp{SOA} \cite{SOAsemantic}, specifying an application they desire and submitting it to their habitat.}{picCapUser} This description enables the definition of a metric for evaluating the \emph{fitness} of a composition of agents, as a distance function between the \emph{semantic description} of the request and the agents' \emph{semantic descriptions}. \setCap{A population is then instantiated in the user's habitat in response to the user's request, seeded from the agents available at their habitat}{picUserReq} (i.e. its agent-pool). This allows the evolutionary optimisation to be accelerated in the following three ways: first, the habitat network provides a subset of the agents available globally, which is localised to the specific user it represents; second, making use of agent-sequences previously evolved in response to the user's earlier requests; and third, taking advantage of relevant agent-sequences evolved elsewhere in response to similar requests by other users. The population then proceeds to evolve the optimal agent-sequence(s) that fulfils the user request, and as the agents are the base unit for evolution, it searches the available agent-sequence combination space. For an evolved agent-sequence that is executed (instantiated) by the user, it then migrates to other peers (habitats) becoming hosted where it is useful, to combine with other agents in other populations to assist in responding to other user requests for applications.

\tfigure{width=3.5in}{userRequestNew}{graffle}{User Request to the Digital Ecosystem}{(modified from \cite{kpic}): A user \getCap{picCapUser} \getCap{picUserReq} (agent-pool).}{-9mm}{}{}{-3mm}

\section{Distributed Intelligence}

\label{tmsection}

Potential exists to optimise the distribution of the agents within the habitat network, through additional \emph{targeted migration} of the agents, which will indirectly optimise the evolving agent populations. The \emph{migration probabilities} between the habitats produces the existing passive agent migration, allowing the agents to spread in the correct general direction within the habitat network, based primarily upon success at their current location. This augmentation will work in a more active manner, allowing the agents highly targeted migration to specific habitats, in addition to the generally directed passive migration. It will help to optimise the agents found at the agent-pools of the habitats, which will in turn optimise the evolving agent populations as they make use of the agent-pools when determining applications (agent-sequences) to user requests. So, accelerating the process of \emph{ecological succession} \cite{begon96}, and therefore the responsiveness of the Digital Ecosystem to the user base.

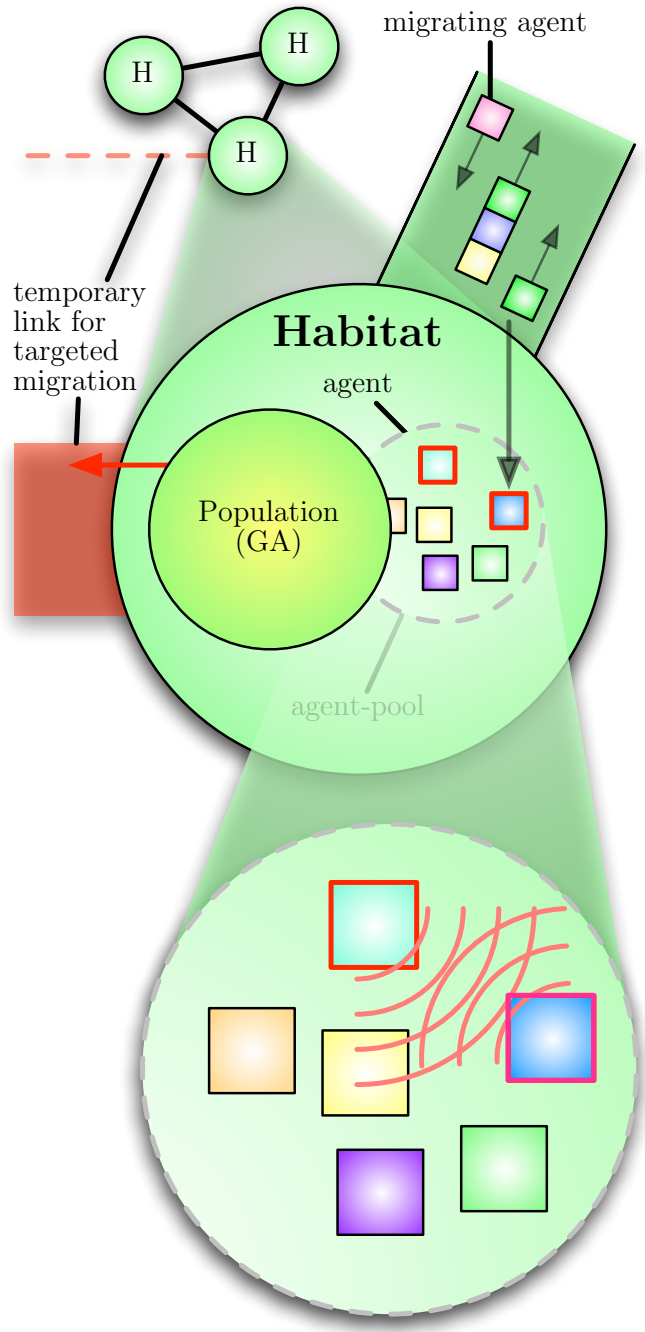
\tfigure{width=3.5in}{complementMIGRATION}{graffle}{Targeted Migration}{This augmentation will optimise the distribution of the agents within the habitat network, through additional {targeted migration} of the agents, helping to {optimise} the agents found at the agent-pools of the habitats, which will in turn optimise the evolving agent populations. So, {accelerating} the process of {ecological succession} \cite{begon96}, and therefore the responsiveness of the Digital Ecosystem to the user base.}{-9mm}{}{}{}

The \emph{targeted migration} will work by providing the agents with the opportunity to interact inside the agent-pools, outside of the evolutionary optimisation of the evolving agent populations, to determine if they are functionally similar based on their \emph{semantic descriptions}. Similar agents will compare their \emph{migration histories} to determine habitats where they could find a niche (i.e. be valuable). This will lead to additional highly targeted migration of agents throughout the habitat network, optimising the set of agents and agent-sequences at the agent-pools, and therefore indirectly optimising and accelerating the evolving agent populations within the Digital Ecosystem. As each evolving agent population within the Digital Ecosystem will be accelerated, the entire ecosystem will operate more efficiently. The \emph{targeted migration} will strengthen the agent concept within the \acl{EOA} of Digital Ecosystems, endowing the individual agents with some intelligence and control over their behaviour. Interestingly, the effectiveness of this augmentation relies on the local interactions of the agents, producing an emergent global optimising effect on the evolving agent populations to accelerate the \emph{ecological succession} of a Digital Ecosystem.

The \emph{targeted migration} will directly optimise the ecological migration, and therefore indirectly complement the evolutionary self-organisation of the evolving agent populations, through the highly \emph{targeted migration} of the agents to their niche habitats. The \emph{migration probabilities} between the habitats produces the existing passive agent migration, allowing the agents to spread in the correct general direction within the habitat network, based primarily upon success at their current location. The \emph{targeted migration} will work in a more active manner, allowing the agents highly targeted migration to specific habitats, based upon their interaction with one another to discover habitats where they could be valuable (i.e. find a niche).

The \emph{targeted migration} will occur when users deploy their services, specifically when deploying their representative agents to their habitats within the Digital Ecosystem, and upon the execution of applications (groups of services), specifically the resulting passive migration of their representative agent-sequences between the habitats. The agent-sequences arriving at habitats, with respect to the \emph{targeted migration}, will be treated as individual agents arriving at the habitats. So, an agent arriving at a habitat interacts one-on-one with agents already present within the agent-pool of the habitat, and upon determining functional similarity, based upon comparing their \emph{semantic descriptions}, will share other habitats successfully visited from their respective \emph{migration histories}. An agent \emph{migration history} is the migratory path of the agent through the habitat network, including its use at the habitats visited. So, similar agents can share their \emph{migration histories} to discover new habitats where they could be valuable, and then use \emph{targeted migration} (via a copy, and not a move) to explore the most promising of the recently acquired habitats. This will allow successfully interacting agents to target specific habitats where they will potentially be useful, but risks potentially infinite \emph{targeted migration}, because \emph{targeted migration} itself can lead to further \emph{targeted migration}. So, each agent will require a dynamic \emph{targeted migrations} counter, which defines the number of permitted targeted migrations of the agent. This counter will be incremented upon an agent's execution in response to a user request, and decremented upon performing a \emph{targeted migration}.

\tfigure{width=3.5in}{newLifeCycle}{graffle}{Agent Life-Cycle With Targeted Migration}{\getCap{nlccap} \getCap{nlc2cap}.}{-6mm}{}{}{}

\setCap{The agent life-cycle will change to support the \emph{targeted migration}, as shown}{nlccap} in Figure \ref{newLifeCycle} \setCap{by the blue circle. Specifically, there will be more opportunities for agent migration, but more importantly these opportunities will be for \emph{targeted migration}, which will help to optimise the set of agents found at the habitats}{nlc2cap}, and therefore support the evolving agent populations created in response to user requests for applications. The \emph{targeted migration} will essentially short-circuit the hierarchical topology habitat network, which is what allows it to specialise and localise to communities, providing specific solutions to specific requests from specific users. However, the \emph{targeted migration} will also reinforce the hierarchical topology of the habitat network, because \emph{targeted migration} between connected habitats will accelerate the existing migration of agents, while between unconnected habitats will assist the Digital Ecosystem in supporting emerging communities. So, the \emph{targeted migration} will help strengthen and catalyse the formation of clusters within the habitat network, and will also assist in locating habitats within the correct clusters. Therefore, the optimisation of the Digital Ecosystem will be a global emergent effect resulting from the local interactions of the agents, allowing for niches to be fulfilled faster and so accelerating the process of \emph{ecological succession} \cite{begon96}. Also, the Digital Ecosystem will adapt faster to changing environmental conditions (e.g. changes in the request behaviour of user communities). In biological terms the \emph{targeted migration} endows the agents with a form of \emph{reciprocal altruistic behaviour} \cite{trivers1971era}, consistent with the agent paradigm of the \acl{EOA}.

\subsection{Similarity Recognition}

For the \emph{targeted migration} to work successfully an effective technique will be required for the \emph{similarity recognition} between the \emph{semantic descriptions} of two agents. Each agent will have an embedded \emph{similarity recognition} component to maintain the consistency of the agent paradigm of \aclp{EOA}. So, the agents will interact one-on-one to determine functional similarity based upon their \emph{semantic descriptions}, using their embedded \emph{similarity recognition} components, with each of the two interacting agents determining similarity for themselves. Again, this is to maintain the consistency of the agent paradigm. \emph{Similarity recognition} between the \emph{semantic descriptions} of two agents will require some form of \emph{pattern recognition}, because there is no single standard for the \emph{semantic description} of services \cite{cabral2004asw}, and we would not adopt any one over the others as it would be inconsistent with the inclusive nature of Digital Ecosystems. So, we will now consider the field of \emph{pattern recognition} to determine suitable techniques for the \emph{similarity recognition} components to be embedded within the agents.

\emph{Pattern recognition} aims to classify data (patterns) based on \emph{priori knowledge} or on statistical information extracted from the data \cite{ripley1996pra}. \emph{Pattern recognition} requires a sensor or sensors for data acquisition, a pre-processing technique, a data representation scheme, and a decision making model \cite{jain2000spr}. Also, learning from a set of examples (training set) is an important and desirable feature of most pattern recognition systems \cite{ripley1996pra}. The four best known approaches for pattern recognition are \cite{jain2000spr}: 
\begin{itemize}
\item Template Matching
\item Statistical Classification
\item Structural Matching
\item Neural Networks
\end{itemize}
These approaches are not necessarily independent, and sometimes the same pattern recognition method exists with different interpretations \cite{jain2000spr}. For example, attempts have been made to design hybrid systems involving multiple approaches, such as the notion of \emph{attributed grammars} which unifies structural and statistical pattern recognition \cite{fu1982spr}.

Template Matching is not suitable for the required \emph{pattern recognition} of our \emph{targeted migration}, because its effective use is domain specific \cite{jain2000spr} and the \emph{similarity recognition} between the \emph{semantic descriptions} of agents is very different to the domains that it is typically applied \cite{nixon2007fei}. Statistical Classification is also not suitable, because the embedded \emph{similarity recognition} component of each agent would require human intervention for variable selection and transformation \cite{michie1995mln}. Structural Matching is suitable theoretically, but implementations lead to many difficulties \cite{jain2000spr}, including the segmentation of noisy patterns (to detect primitives) and the inference of grammar from training data \cite{jain2000spr}. There can also be a combinatorial explosion of possibilities to be investigated, demanding large training sets and significant computational effort \cite{perlovsky1998ccc}, neither of which is available. \aclp{NN} are suitable, given their low dependence on domain-specific knowledge and the availability of efficient learning algorithms \cite{jain2000spr}. \aclp{SVM}, albeit a recent development \cite{vapnik1998slt}, are also suitable \cite{joachims1997tcs}, being primarily a binary classifier \cite{jain2000spr} for training generalisable nonlinear classifiers in high-dimensional spaces using small training sets \cite{valafar2002s}. So, we will make use of both \aclp{NN} and \aclp{SVM} for our \emph{targeted migration}.

\subsubsection{Neural Networks}

In the first instance, we will leverage the \emph{pattern recognition} capabilities of \acfp{NN} for the embedded \emph{similarity recognition} components of the agents, allowing them to determine similarity to one another based on the similarity of their \emph{semantic descriptions}. We will use \emph{multilayer perceptrons} (feed-forward artificial \acp{NN}) with \emph{backpropagation} \cite{haykin1998nnc} to provide the required \emph{pattern recognition} behaviour, because of their ability to solve problems stochastically, which allows for approximate solutions to extremely complex problems \cite{haykin1998nnc}. They are a modification of the \emph{standard linear perceptron} \cite{rojas1996nns}, using three or more layers of neurons (nodes) with nonlinear activation functions to distinguish data that is not linearly separable, or separable by a hyperplane \cite{bishop1995nnp}. The power of the multilayer perceptron comes from its similarity to certain biological neural networks in the human brain, and because of their wide applicability has become the standard algorithm for any supervised-learning \emph{pattern recognition} process \cite{haykin1998nnc}.

\acp{NN} can be viewed as massively parallel computing systems consisting of an extremely large number of simple processors with many interconnections \cite{ripley1996pra}. \ac{NN} models attempt to use certain organisational principles (such as learning, generalisation, adaptivity, fault tolerance, distributed representation, and computation) in a network of weighted directed graphs, in which the nodes are artificial neurons, and the directed edges (with weights) are connections between the neuron outputs and inputs \cite{ripley1996pra}. The main characteristics of \acp{NN} are their ability to learn complex nonlinear input-output relationships, use sequential training procedures, and adapt themselves to the data \cite{jain2000spr}. The most commonly used family of \acp{NN} for pattern classification tasks is the feed-forward network, including multilayer perceptrons, which are organised into layers and has unidirectional connections between the layers \cite{jain2000spr}. Another popular network is the \acl{SOM}, or Kohonen-Network \cite{kohonen}, which is often used for feature mapping \cite{jain2000spr}. The increasing popularity of NN models to solve pattern recognition problems has been primarily because of their low dependence on domain-specific knowledge (relative to model-based and rule-based approaches) and the availability of efficient learning algorithms \cite{jain2000spr}. The learning process involves updating the network architecture and connection weights so that a network can efficiently perform a specific classification \cite{ripley1996pra}. \acp{NN} provide a suite of nonlinear algorithms for feature extraction (using hidden layers) and classification (e.g. multilayer perceptrons) \cite{jain2000spr}. In addition, existing feature extraction and classification algorithms can be mapped onto \ac{NN} architectures for efficient (hardware) implementation \cite{table1994hia}. Despite the seemingly different underlying principles, most of the well-known \ac{NN} models are implicitly equivalent or similar to classical statistical pattern recognition methods \cite{jain2000spr}. However, \acp{NN} offer several advantages, such as unified approaches for feature extraction and classification, and flexible procedures for finding good, moderately nonlinear solutions \cite{jain2000spr}. 

\label{NNsubsection}

A pre-processing \cite{bishop1995nnp} of agent \emph{semantic descriptions} will be required that is consistent across the entire Digital Ecosystem, requiring an alphabetical ordering of the attribute tuples within a \emph{semantic description}, a standardisation of the length of the attributes, before finally making use of a binary encoding for processing by a \ac{NN} \cite{bishop1995nnp}. The assumption of information structured as tuples, including an attribute name and attribute value, is accurate for our simulated \emph{semantic descriptions}, but is also a reasonable assumption for any \emph{semantic description} of web services \cite{dustdar2005sws, medjahed2003cws, debruijn2006wsm, cabral2004asw}. To standardise the length of the attributes, after removing any white-space\footnote{A white-space is any single character or series of characters that represent horizontal or vertical space in typography \cite{friedl2006mre}.}, an average word length of six characters will be used, because 5.39 is the average word length for business English \cite{fox1999sim}. For the binary encoding we propose using Unicode (UTF-8), which is based on extending ASCII to provide multilingual support \cite{gillam2002udp}. However, ASCII's support of only English \cite{gillam2002udp} will be sufficient for our simulations. The size (number of neurons) of the \emph{input layer} \cite{haykin1998nnc} will be proportional to the \emph{semantic description} of the agent in which the \ac{NN} is embedded, taking advantage of the variation in length of different \emph{semantic descriptions}, which will assist the \ac{NN}-based \emph{pattern recognition} in determining dissimilarity.

We will use a single \emph{hidden layer}, which is usually sufficient for most tasks \cite{bishop1995nnp}. The size of which will be determined through \emph{exploratory programming} \cite{sommerville2006se} in our simulations, because of the difficulty in determining the optimal size without training several networks and estimating the generalisation error \cite{sarle2008}, evident by the range of inconsistent \emph{rules of thumb} \cite{blum1992nnc, swingler1996, berry1997dmt, boger1997kea} available to define the optimal size. The \emph{output layer} \cite{bishop1995nnp} will consist of a single neurone to provide a binary (true or false) response to the question of whether another agent's \emph{semantic description} is similar to the agent's own \emph{semantic description}. We will use a threshold of $0.90$ on its output for the determination of similarity. The overall structure of the \acl{NN} is visualised in Figure \ref{NNstructure}.

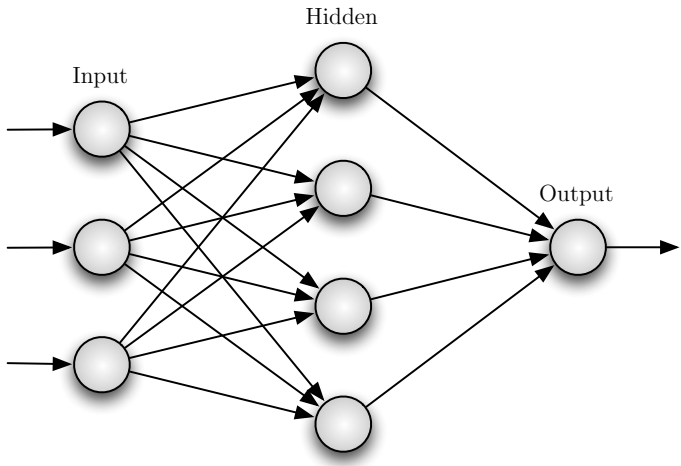
\tfigure{width=3.5in}{NNstructure}{graffle}{Neural Network for the Similarity Recognition Component}{of agents in Targeted Migration: Consisting of an {input layer} proportional to the {semantic description} of the agent in which it is embedded. A single {hidden layer}, and an {output layer} consisting of a single neurone to provide a binary response to the question of whether another agent's {semantic description} is similar.}{-9mm}{}{}{}

\emph{Multilayer perceptrons} use nonlinear activation functions, which were developed to model the frequency of action potentials (firing) of biological neurons in the brain \cite{haykin1998nnc}. The main activation function used in current applications is the sigmoid function \cite{haykin1998nnc}, a hyperbolic tangent that is normalised and in which the output $y$ of a neurone is the sum of the weighted input values $x$ \cite{bishop1995nnp}, 
\begin{equation}
y = \frac{1}{(1 + e^{-x})}.
\label{sigmoid}
\end{equation}
The weights $x$ between the neurons will be randomly initialised, then trained to the real numbers that provide the desired functionality, because learning occurs in the perceptron by changing the connection (synaptic) weights after each piece of data is processed, based on the error of the output compared to the expected result \cite{bishop1995nnp}. This is an example of \emph{supervised learning} and is carried out through backpropagation, a generalisation of the least mean squares algorithm \cite{haykin1998nnc}. The network is therefore trained by providing it with input and corresponding output patterns \cite{bishop1995nnp}.

The \ac{NN}-based embedded \emph{similarity recognition} component of an agent will be trained when the agent is deployed to a habitat of the Digital Ecosystem. The initial \emph{training set} will consist of the \emph{semantic description} of the agent as a positive match, and variants created from its own \emph{semantic description}. If the variant is less than 10\% different it will be processed as a positive match, else it will be processed as a negative match. The \emph{training set} can be extended based on experience, making use of when an agent visits a habitat through \emph{targeted migration} (i.e. one acquired from an inter-agent interaction); if visiting the habitat proves successful the \emph{semantic description} of the interacting agent can be appended to the \emph{training set} as a positive match, else as a negative match.

\subsubsection{Support Vector Machines}

\label{SVMsubsection}

In the second instance, we will leverage the \emph{pattern recognition} capabilities of \acfp{SVM} for the embedded \emph{similarity recognition} components of the agents, allowing them to determine similarity to one another based on the similarity of their \emph{semantic descriptions}. As \acp{SVM} are closely related to \aclp{NN}, being a close cousin to classical multilayer perceptrons \cite{abe2005svm}, we will make use of the \emph{pre-processing} and the \emph{training sets} defined in the previous subsection, which will also ensure a fair comparison of the \emph{pattern recognition} techniques in empowering the \emph{similarity recognition} components of the agents.

One of the most interesting recent developments in classifier design is the introduction of the \ac{SVM} \cite{vapnik1998slt}, which is primarily a two-class classifier, and therefore highly suitable for the required \emph{similarity recognition} component of our \emph{targeted migration}. It uses an optimisation criterion that is the width of the margin between the classes, i.e. the empty area around the decision boundary defined by the distance to the nearest training patterns \cite{burges98tutorial}. These patterns, called \emph{support vectors}, define the classification function, and their number is minimised by maximising the margin \cite{burges98tutorial}. This is achieved through a kernel function $K$, which transposes the data into a higher-dimensional space where a hyperplane performs the separation \cite{burges98tutorial}. In its simplest form the kernel function is just a dot product between the input pattern and a member of the support set, resulting in a linear classifier, while nonlinear kernel functions lead to a polynomial classifier \cite{jain2000spr}. \acp{SVM} are closely related to \aclp{NN}, being a close cousin to classical multilayer perceptrons, with the use of a sigmoid kernel function making them equivalent to two-layer perceptrons \cite{abe2005svm}. However, in the training of \acp{NN}, such as multi-layer perceptrons, the weights of the network are found by solving a non-convex unconstrained minimisation problem, while the use of a kernel function in \acp{SVM} solves a quadratic programming problem with linear constraints \cite{scholkopf1997csv}. An important advantage of \acp{SVM} is that they offer the possibility to train generalisable nonlinear classifiers in high-dimensional spaces using a small training set \cite{valafar2002s}. Furthermore, for large training sets a small support set is typically selected for designing the classifier, thereby minimising the computational requirements during training \cite{valafar2002s}.

The selection of a suitable kernel function is important, since it defines the \emph{feature space} in which the \emph{training set} is classified \cite{svm6}, operating as shown in Figure \ref{svm}. \setCap{A \acf{RBF} is recommended for text categorisation \cite{joachims1997tcs}, with the most common form of the \ac{RBF} being Gaussian \cite{gunn1998svm}.}{svmCap}

\tfigure{width=3.5in}{svm}{graffle}{\acl{SVM}}{(modified from \cite{takahashi2008}): Visualisation showing the training set in the Input Space, and its binary classification by a hyperplane in the higher dimensional Feature Space, achieved through the kernel function. \getCap{svmCap}}{-6mm}{}{}{}

Training a \ac{SVM} requires solving a large \ac{QP} optimisation problem, which \acf{SMO} breaks into a series of the smallest possible \ac{QP} problems. \ac{SMO} solves these small \ac{QP} problems analytically, which avoids using a time-consuming numerical \ac{QP} optimisation. \ac{SMO} scales between linear and quadratic time complexity, relative to the size of the \emph{training set}, because it avoids matrix computation \cite{platt1999fts}. The alternative, a standard \ac{PCG} chunking algorithm scales between linear and cubic time complexity, relative to the size of the \emph{training set} \cite{platt1999fts}. So \ac{SMO} is faster, up to a thousand times on real-world sparse data sets \cite{platt1999fts}.

The issue of the learnt behaviour of the embedded \emph{similarity recognition} component of an agent, whether \ac{SVM} or \ac{NN} based, being inherited when the agent reproduces is known as the Baldwin effect \cite{baldwin1896nfe}. The Baldwin effect has always been controversial within biological ecosystems \cite{weber2003eal}, primarily because of the problem of confirming it experimentally \cite{sterelny2004rea}. Also, offspring in \emph{biological ecosystems} can be genetically different to their parents \cite{begon96}, such that any learnt behaviour could potentially be inappropriate. However, the offspring in our Digital Ecosystem are genetically identical to their parents (in terms of the individual agents), and so it makes little sense to force the loss of learnt behaviour. So, we doubt that the Baldwin effect will adversely affect our Digital Ecosystem, which we will confirm through our simulations.

Now that the \emph{targeted migration} is theoretically complete, with two alternative \emph{pattern recognition} techniques, we can confirm its effect experimentally through simulations.

\section{Simulation and Results}

Although agent-based modelling solutions, like Repast (Recursive Porous Agent Simulation Toolkit) \cite{collier2003ref} and MASON (Multi-Agent Simulator Of Neighbourhoods) \cite{luke2004mnm}, and evolutionary computing libraries, like ECJ (Evolutionary Computing in Java) \cite{ECJ} and the JCLEC (Java Computing Library for Evolutionary Computing) \cite{ventura2008jjf}, are available, it was evident that it would take as much effort to adapt one, or a combination, of these to simulate the Digital Ecosystem, as it would to create our own simulation of the Digital Ecosystem, because the required ecological dynamics are largely absent from these and other available technologies. So, we created our own simulation, following the \acl{EOA} from section \ref{simRes}, using the \emph{business ecosystem} of \aclp{SME} from \aclp{DBE} \cite{dbebkintro} as an example user base, adding the classes and methods necessary to implement the proposed \emph{targeted migration} augmentation. Each experimental scenario was run ten thousand times for statistical significance of the means and standard deviations calculated. We implemented the \emph{targeted migration} as defined in section \ref{tmsection}, using both \acf{NN} and \acf{SVM} based \emph{similarity recognition} components embedded within the agents.

\subsection{Agents: Semantic Descriptions}
\label{descriptions}

\mfigure{\begin{center}\bold{A = \{(1,25), (2,35), (3,55), (4,6), (5,37), (6,12)\}}\end{center}}{Agent Semantic Descriptions}{\getCap{as4} \getCap{as3} between three and six numeric tuples; each \getCap{agentSemantic2}}{BMLprocess}{0mm}{-3mm}{!h}

An agent represents a user's service, including the \emph{semantic description} of the \emph{business process} involved, and is based on existing and emerging technologies for \emph{semantically capable} \aclp{SOA} \cite{SOAsemantic}, such as the \acs{OWL-S} semantic markup for web services \cite{martin2004bsw}. We simulated a service's \emph{semantic description} \setCap{with an abstract representation consisting of a set of}{as3} numeric tuples, to simulate the properties of a \emph{semantic description}. Each \setCap{tuple representing an \emph{attribute} of the \emph{semantic description}, one integer for the \emph{attribute identifier} and one for the \emph{attribute value}, with both ranging between one and a hundred.}{agentSemantic2} \setCap{Each simulated agent had a semantic description}{as4}, with between three and six tuples, as shown in Figure \ref{BMLprocess}.

\subsection{User Base}

\mfigure{\begin{center}\bold{R = [\{(1,23),(2,45),(3,33),(4,6),(5,8),(6,16)\}, \{(1,84),(2,48),(3,53),(4,11),(5,16)\}]}\end{center}}{User Request}{\getCap{semanticRequest}; each \getCap{agentSemantic2}}{userRequest}{0mm}{-3mm}{!h}

Throughout the simulations we assumed a hundred users, which meant that at any time the number of users joining the network equalled those leaving. The habitats of the users were randomly connected at the start, to simulate the users going online for the first time. The users then produced agents (services) and requests for business applications. Initially, the users each deployed five agents to their habitats, for migration (distribution) to any habitats connected to theirs (i.e. their community within the \emph{business ecosystem}). Users were simulated to deploy a new agent after the submission of three requests for business applications, and were chosen at random to submit their requests. \setCap{A simulated user request consisted of an abstract \emph{semantic description}, as a list of sets of numeric tuples to represent the properties of a desired business application}{semanticRequest}. The use of the \emph{numeric tuples} made it comparable to the \emph{semantic descriptions} of the services represented by the agents; while the \emph{list of sets} (two level hierarchy) and a much longer length provided sufficient complexity to support the sophistication of business applications. An example is shown in Figure \ref{userRequest}.

The user requests were handled by the habitats instantiating evolving populations, which used evolutionary computing to find the optimal solution(s), agent-sequence(s). It was assumed that the users made their requests for business applications \emph{accurately}, and always used the response (agent-sequence) provided.

\subsection{Populations: Evolution}
\label{fitnessFunction}

Populations of agents, $[A_1, A_1, A_2, ...]$, were evolved to solve user requests, seeded with agents and agent-sequences from the \emph{agent-pool} of the habitats in which they were instantiated. A dynamic population size was used to ensure exploration of the available combinatorial search space, which increased with the average length of the population's agent-sequences. The optimal combination of agents (agent-sequence) was evolved to the user request $R$, by an artificial \emph{selection pressure} created by a \emph{fitness function} generated from the user request $R$. An individual (agent-sequence) of the population consisted of a set of attributes, ${a_1, a_2, ...}$, and a user request essentially consisted of a set of required attributes, ${r_1, r_2, ...}$. So, the \emph{fitness function} for evaluating an individual agent-sequence $A$, relative to a user request $R$, was
\begin{equation}
fitness(A,R) = \frac{1}{1 + \sum_{r \in R}{|r-a|}},
\label{ff}
\end{equation}
where $a$ is the member of $A$ such that the difference to the required attribute $r$ was minimised. Equation \ref{ff} was used to assign \emph{fitness} values between 0.0 and 1.0 to each individual of the current generation of the population, directly affecting their ability to replicate into the next generation. The evolutionary computing process was encoded with a low mutation rate, a fixed selection pressure and a non-trapping fitness function (i.e. did not get trapped at local optima). The type of selection used \emph{fitness-proportional} and \emph{non-elitist}, \emph{fitness-proportional meaning that the \emph{fitter} the individual the higher its probability of} surviving to the next generation \cite{blickle1996css}. \emph{Non-elitist} means that the best individual from one generation was not guaranteed to survive to the next generation; it had a high probability of surviving into the next generation, but it was not guaranteed as it might have been mutated \cite{eiben2003iec}. \emph{Crossover} (recombination) was then applied to a randomly chosen 10\% of the surviving population, a \emph{one-point crossover}, by aligning two parent individuals and picking a random point along their length, and at that point exchanging their tails to create two offspring \cite{eiben2003iec}. \emph{Mutations} were then applied to a randomly chosen 10\% of the surviving population; \emph{point mutations} were randomly located, consisting of \emph{insertions} (an agent was inserted into an agent-sequence), \emph{replacements} (an agent was replaced in an agent-sequence), and \emph{deletions} (an agent was deleted from an agent-sequence) \cite{lawrence1989hsd}. The issue of bloat was controlled by augmenting the \emph{fitness function} with a \emph{parsimony pressure} \cite{soule1998ecg} which biased the search to shorter agent-sequences, evaluating longer than average length agent-sequences with a reduced \emph{fitness}, and thereby providing a dynamic control limit which adapted to the average length of the ever-changing evolving agent populations.

\subsection{Semantic Filter}
\label{semanticFilter}

\mfigure{\begin{center}Agent's semantic description:\end{center}
\vspace{-2mm}
\begin{quote}\begin{center}\bold{\{\red{(1,25)}, \blue{(2,35)}, \yellow{(3,55)}, \green{(4,6)}, \purple{(5,37)}, \turquoise{(6,12)}\}}\end{center}
\end{quote}
\begin{center}(with semantic filter):\end{center}
\vspace{-2mm}
\begin{quote}\begin{center}
\bold{\{\red{(Business, Airline)}, \blue{(Company, British Midland)}, \yellow{(Quality, Economy)}, \green{(Cost, 60)}, \purple{(Depart, Edinburgh)}, \turquoise{(Arrive, London)}\}}\end{center}\end{quote}
\begin{center}user request:\end{center}
\vspace{-2mm}
\begin{quote}\bold{[\{\red{(1,23)}, \blue{(2,45)}, \yellow{(3,33)}, \green{(4,6)}, \purple{(5,8)}, \turquoise{(6,16)}\}, \{\red{(1,84)}, \blue{(2,48)}, \yellow{(3,53)}, \green{(4,11)}, \grey{(7,16)}, \brown{(8,34)}\}, \{\red{(1,23)}, \blue{(2,45)}, \yellow{(3,53)}, \green{(4,6)}, \purple{(5,16)}\turquoise{(6,53)}\}, \{\red{(1,86)}, \blue{(2,48)}, \yellow{(3,33)}, \green{(4,25)}, \grey{(7,55)}\brown{(8,23)}\}, \{\red{(1,25)}, \blue{(2,52)}, \yellow{(3,53)}, \green{(4,5)}, \purple{(5,55)}, \turquoise{(6,37)}\}, \{\red{(1,86)}, \blue{(2,48)}, \yellow{(3,43)}, \green{(4,25)}, \grey{(7,37)}, \brown{(8,40)}\}, \{\red{(1,22)}, \blue{(2,77)}, \yellow{(3,82)}, \green{(4,9)}, \purple{(5,35)}, \turquoise{(6,8)}\}]}
\end{quote}
\begin{center}(with semantic filter):\end{center}
\vspace{-2mm}
\begin{quote}
\bold{[\{\red{(Business, Airline)}, \blue{(Company, Air France)}, \yellow{(Quality, Economy)}, \green{(Cost, 60)}, \purple{(Depart, Edinburgh)}, \turquoise{(Arrive, Paris)}\}, \{\red{(Business, Hotel)}, \blue{(Company, Continental)}, \yellow{(Quality, 3*)}, \green{(Cost, 110)}, \grey{(Location, Paris)}, \brown{(Nights, 3)}\}, \{\red{(Business, Airline)}, \blue{(Company, Air France)}, \yellow{(Quality, Economy)},\green{(Cost,60)},\purple{(Depart, Paris)}, \turquoise{(Arrive, Monte Carlo)}\}, \{\red{(Business, Hotel)}, \blue{(Company, Continental)}, \yellow{(Quality, 2*)}, \green{(Cost, 250)}, \grey{(Location, Monte Carlo)}, \brown{(Nights, 2)}\}, \{\red{(Business, Airline)}, \blue{(Company, KLM)}, \yellow{(Quality, Economy)}, \green{(Cost, 50)}, \purple{(Depart, Monte Carlo)}, \turquoise{(Arrive, London)}\}, \{\red{(Business, Hotel)}, \blue{(Company, Continental)}, \yellow{(Quality, 3*)}, \green{(Cost, 250)}, \grey{(Location, London)}, \brown{(Nights, 4)}\}, \{\red{(Business, Airline)}, \blue{(Company, Air Espana)}, \yellow{(Quality, First)}, \green{(Cost, 90)}, \purple{(Depart, London)}, \turquoise{(Arrive, Edinburgh)}\}]}
\end{quote}}{Semantic Filter}{Shows \getCap{bmlcap1} \getCap{capbml2}. \getCap{capbml3}}{BMLreal}{0mm}{-2mm}{!h}

The simulation of the Digital Ecosystem complies with the \acl{EOA} defined in the previous section, but there was the possibility of model error in the \emph{business ecosystems} of the user base (\aclp{SME} from \aclp{DBE} \cite{dbebkintro}), because while the abstract numerical definition for the simulated \emph{semantic descriptions}, of the services and requests the users provide, makes it widely applicable, it was unclear that it could accurately represent \emph{business services}. So we created a \emph{semantic filter} to show \setCap{the \emph{numerical semantic descriptions}, of the simulated services (agents) and user requests, in a \emph{human readable form}.}{bmlcap1} The basic properties of any \emph{business process} are cost, quality, and time \cite{davenport1990nie}; so this was followed in the \emph{semantic filter}. \setCap{The \emph{semantic filter} translates \emph{numerical semantic descriptions} for one community within the user base, showing it in the context of the travel industry}{capbml2}, as shown in Figure \ref{BMLreal}. \setCap{The simulation still operated on the numerical representation for operational efficiency, but the \emph{semantic filter} essentially assigns meaning to the numbers.}{capbml3} The output from the \emph{semantic filter}, in Figure \ref{BMLreal}, shows that the \emph{numerical semantic descriptions} are a reasonable modelling assumption that abstracts sufficiently rich textual descriptions of \emph{business services}.

\subsection{Controls}

The \emph{targeted migration} was dependent on additional agent migration, which alone could have been responsible for any observed optimisation, because it led to greater distribution of the agents within the Digital Ecosystem, potentially improving responsiveness for the user base. So, we included a \emph{migration control} in our experimental simulations for the additional agent migration, being random instead of targeted. Furthermore, to determine the contribution of the \acp{NN} and \acp{SVM} on the \emph{targeted migration} we created a \emph{pattern recognition control}, using a rudimentary \emph{distance function} adapted from our simulated \emph{fitness function} defined in section \ref{fitnessFunction}.

\tfigure{width=3.5in}{graphTarMigContHist}{graph}{Graph of the Targeted Migration Controls and the Digital Ecosystem}{\getCap{tmcf} response rate, \getCap{tmc2f}, \getCap{tmc3f}.}{-7mm}{}{}{}

\tfigure{width=3.5in}{graphTarMigControl}{graph}{Graph of Typical Runs for the Targeted Migration Controls}{and the Digital Ecosystem: \getCap{boohoo}. The migration control with additional random migration ultimately decreased the responsiveness, while the pattern recognition control performed only slightly better.}{-7mm}{}{}{}

In Figure \ref{graphTarMigContHist} we graphed for the simulation runs the average of the percentage response rate after a thousand time steps (user request events), for the Digital Ecosystem with the \emph{migration control}, and the Digital Ecosystem with the \emph{pattern recognition control}, compared to the Digital Ecosystem alone. \setCap{The Digital Ecosystem alone averaged a 68.0\%}{tmcf} (3 s.f.) response rate with a standard deviation of 2.61 (2 d.p.), \setCap{while the Digital Ecosystem with the \emph{migration control} showed a significant degradation to 49.6\%}{tmc2f} (3 s.f.) with a standard deviation of 1.96 (2 d.p.), \setCap{and the Digital Ecosystem with the \emph{pattern recognition control} showed only a small increase to 70.5\%}{tmc3f} (3 s.f.) with a standard deviation of 2.60 (2 d.p.). Therefore, any observed improvement from the \emph{targeted migration} was not from the additional migration but its targeting, and that the effectiveness of the \emph{pattern recognition} functionality will be significant if the \emph{targeted migration} is to be effective.

In Figure \ref{graphTarMigControl} we graphed a typical run of the Digital Ecosystem with the \emph{migration control}, and the Digital Ecosystem with the \emph{pattern recognition control}, compared to the Digital Ecosystem alone \cite{de07oz}. \setCap{The Digital Ecosystem alone performed as expected, adapting and improving over time to reach a mature state}{boohoo} through the process of \emph{ecological succession} \cite{begon96}. The Digital Ecosystem with the \emph{migration control}, which included additional random migration, while initially beneficial, ultimately decreased the responsiveness of the Digital Ecosystem. Finally, the Digital Ecosystem with the \emph{pattern recognition control} performed only marginally better than the Digital Ecosystem alone.

\subsection{Neural Networks}

We started with the \ac{NN}-based \emph{targeted migration}, as defined in section \ref{NNsubsection}. We made use of Joone (Java Object Oriented Neural Engine) \cite{joone} to implement the required \acp{NN}, and \emph{exploratory programming} \cite{sommerville2006se} to determine that a \emph{hidden layer} 1.5 times the size of the \emph{input layer} was effective for the \ac{NN}-based \emph{similarity recognition} components. 

In Figure \ref{graphTarMigHistSVM} we graphed for the simulation runs the average of the percentage response rate after a thousand time steps (user request events), for the Digital Ecosystem with the \ac{NN}-based \emph{targeted migration}, compared to the Digital Ecosystem alone. \setCap{The Digital Ecosystem alone averaged a 68.0\% (3 s.f.) response rate with a standard deviation of 2.61 (2 d.p.), while the Digital Ecosystem with the \ac{NN}-based \emph{targeted migration} showed a significant improvement to a 92.1\% (3 s.f.) response rate with a standard deviation of 2.22 (2 d.p.).}{nnbtmCap}

\tfigure{width=3.5in}{graphTarMigHistSVM}{graph}{Graph of Neural Network and \acl{SVM} Based Targeted Migration}{\getCap{tmsf}, \getCap{tms2f}, \getCap{tms3f}.}{-7mm}{}{}{}

\tfigure{width=3.5in}{graphTarMigSVM}{graph}{Graph of Typical Runs for the Digital Ecosystem and Targeted Migration}{\getCap{gNN2cap} In comparison, the \getCap{gNN3cap}.}{-7mm}{}{}{}

\subsection{Support Vector Machines}

Next we considered the \ac{SVM}-based \emph{targeted migration}, as defined in section \ref{SVMsubsection}, making use of LIBSVM (Library for Support Vector Machines) \cite{libsvm} to implement the required \acp{SVM}. In Figure \ref{graphTarMigHistSVM} we graphed for the simulation runs the average of the percentage response rate after a thousand time steps (user request events), for the Digital Ecosystem with the \ac{SVM}-based \emph{targeted migration}, compared to the Digital Ecosystem with the \ac{NN}-based \emph{targeted migration}, and the Digital Ecosystem alone. \setCap{The Digital Ecosystem with the \ac{SVM}-based \emph{targeted migration} averaged a 92.8\% (3 s.f.) response rate}{tmsf} with a standard deviation of 2.09 (2 d.p.), \setCap{slightly better than the \ac{NN}-based \emph{targeted migration} at 92.1\% (3 s.f.)}{tms2f} with a standard deviation of 2.22 (2 d.p.), \setCap{and so significantly better than the Digital Ecosystem alone at 68.0\% (3 s.f.)}{tms3f} with a standard deviation of 2.61 (2 d.p.).

In Figure \ref{graphTarMigSVM} we graphed typical runs of the \setCap{Digital Ecosystem with the \ac{SVM}-based \emph{targeted migration}, the Digital Ecosystem with the \ac{NN}-based \emph{targeted migration}, and the Digital Ecosystem alone.}{gSVM2cap} \setCap{The Digital Ecosystem alone performed as expected, adapting and improving over time to reach a mature state through the process of \emph{ecological succession} \cite{begon96}}{gNN2cap}, approaching 70\% effectiveness for the user base. The \setCap{Digital Ecosystem with the \emph{targeted migration}, \ac{NN} or \ac{SVM}-based, showed a significant improvement}{gNN3cap} in the \emph{ecological succession}, reaching the same performance in less than a fifth of the time, before reaching over 90\% effectiveness for the user base. To show more clearly \setCap{the greater effectiveness of the SVM-based \emph{targeted migration}, compared to the NN-based \emph{targeted migration}}{histogram2Cap}, we graphed in Figure \ref{NNhistogram} the \setCap{frequency of poor matches ($<$50\%) every one hundred time steps, for the Digital Ecosystem with the SVM-based \emph{targeted migration}, compared to the Digital Ecosystem with the NN-based \emph{targeted migration}, and the Digital Ecosystem alone}{histogramCap}.

\tfigure{}{NNhistogram}{graph}{Graph of Frequencies for the Targeted Migration}{The \getCap{histogramCap}. It shows \getCap{histogram2Cap} from the seven hundredth generation onwards.}{-3mm}{}{}{-3mm}

\section{Conclusion}

The results showed that the \emph{targeted migration} optimised and accelerated the \emph{ecological succession} \cite{begon96} of our Digital Ecosystem, constructively interacting with its ecological and evolutionary dynamics. The results also showed that it was not the additional migration, but its targeting that created the improvement in the Digital Ecosystem, and that an effective \emph{pattern recognition} technique was required for the \emph{targeted migration} to operate effectively. Both \acp{NN} and \acp{SVM} proved to be effective, \acp{SVM} marginally more than \acp{NN}. The results also showed that there were no adverse side-effects from the Baldwin effect \cite{baldwin1896nfe}, the inheritance of learnt behaviour in the agents from the embedded \emph{similarity recognition} components, whether \ac{SVM} or \ac{NN} based. Finally, based on the experimental results, and our theoretical understanding, we would recommend \acp{SVM} for the \emph{pattern recognition} functionality of the \emph{targeted migration}. 

A partial reference implementation \cite{eveNet} for our Digital Ecosystem, which includes an implementation of the \emph{targeted migration}, was created by the \acf{DBE} project \cite{DBE}, and we expect that once completed will be deployed as part of the software platform intended for the regional deployment of their \emph{Digital Ecosystems} \cite{den4dek, OPAALS}. Also, an open-source simulation framework for Digital Ecosystems \cite{eveSim} was created by the \aclp{DBE} project \cite{DBE}, and is currently supported by the \acf{OPAALS} project \cite{OPAALS} to assist further research into Digital Ecosystems, including the wider implications of interacting with social systems, such as \emph{business ecosystems} of \acfp{SME}. This will provide the opportunity to investigate further the \emph{complex system} that the Digital Ecosystem represents, and eventually collect real world data to determine whether the Digital Ecosystem can perform usefully in a natural setting.

\bibliographystyle{IEEEtran.bst}
\bibliography{../../../../../../../Desktop/PhDthesis/references}

\end{document}

%% file: acronyms.tex
\acrodef{PCG}{Projected Conjugate Gradient} 
\acrodef{QP}{quadratic programming}
\acrodef{RBF}{Radial-Basis Function}
\acrodef{ABM}{Agent-Based Modelling}
\acrodef{AI}{Artificial Intelligence}
\acrodef{DAI}{Distributed Artificial Intelligence}
\acrodef{API}{Application Programming Interface}
\acrodef{ARF}{p14ARF human tumor-suppressor gene}
\acrodef{B2B}{business-to-business}
\acrodef{BDP}{Biological Design Pattern}
\acrodef{BGS}{Best Guess Solution}
\acrodef{BIC}{Biologically-Inspired Computing}
\acrodef{BML}{Business Modelling Language}
\acrodef{BPEL}{Business Process Execution Language}
\acrodef{BPMN}{Business Process Modelling Notation}
\acrodef{CAS}{Complex Adaptive Systems}
\acrodef{COBOL}{COmmon Business-Oriented Language}
\acrodef{DBE}{Digital Business Ecosystem}
\acrodef{DE}{Digital Ecosystem}
\acrodef{DEC}{distributed evolutionary computing}
\acrodef{DGA}{Distributed genetic algorithms}
\acrodef{DIS}{Distributed Intelligence System}
\acrodef{DNA}{Deoxyribose Nucleic Acid}
\acrodef{DOP}{DBE Open Protocol}
\acrodef{DSS}{Distributed Storage System}
\acrodef{EAP}{Evolving Agent Population}
\acrodef{ebXML}{e-business eXtensible Markup Language}
\acrodef{EC}{Evolutionary Computing}
\acrodef{ECJ}{Evolutionary Computing in Java}
\acrodef{EE}{Evolutionary Environment}
\acrodef{EFL}{Evolutionary Framework for Language}
\acrodef{FLE}{Framework for Language Ecosystems}
\acrodef{EOA}{Ecosystem-Oriented Architecture}
\acrodef{ESS}{evolutionary stable strategy}
\acrodef{EvE}{Evolutionary Environment}
\acrodef{ExE}{Execution Environment}
\acrodef{FCB}{Framework for Computational Biomimicry}
\acrodef{FFF}{Fitness Function Framework}
\acrodef{FL}{Fitness Landscape}
\acrodef{HWU}{Heriot-Watt University}
\acrodef{ICL}{Imperial College London}
\acrodef{ICT}{Information and Communications Technology}
\acrodef{INTEL}{Intel Ireland}
\acrodef{IPA}{International Phonetic Alphabet}
\acrodef{ISUFI}{Istituto Superiore Universitario di Formazione Interdisciplinare}
\acrodef{JDJ}{Java Developer's Journal}
\acrodef{KC}{Kolmogorov-Chaitin}
\acrodef{LAN}{local area network}
\acrodef{LSE}{London School of Economics and Political Science}
\acrodef{MAS}{Multi-Agent System}
\acrodef{MDL}{Minimum Description Length}
\acrodef{MDM2}{murine double minute 2}
\acrodef{MFT}{Mean Field Theory}
\acrodef{MoAS}{Mobile Agent System}
\acrodef{MOF}{Meta Object Facility}
\acrodef{MUH}{migration and usage history}
\acrodef{NIC}{Nature Inspired Computing}
\acrodef{NN}{Neural Network}
\acrodef{NoE}{Network of Excellence}
\acrodef{OMG}{Open Mac Grid}
\acrodef{OPAALS}{Open Philosophies for Associative Autopoietic Digital Ecosystems}
\acrodef{P2P}{peer-to-peer}
\acrodef{P53}{protein 53}
\acrodef{PDA}{Personal Digital Assistant}
\acrodef{QoS}{quality of service}
\acrodef{REST}{REpresentational State Transfer}
\acrodef{RNA}{Deoxyribose Nucleic Acid}
\acrodef{SAE}{Software Agent Ecosystem}
\acrodef{SBML}{Systems Biology Modelling Language}
\acrodef{SBVR}{Semantics of Business Vocabulary and Business Rules}
\acrodef{SDL}{Service Description Language}
\acrodef{SF}{Service Factory}
\acrodef{SIM}{Social Interaction Mechanism}
\acrodef{SM}{Service Manifest}
\acrodef{SME}{Small and Medium sized Enterprise}
\acrodef{SML}{Service Modelling Language}
\acrodef{SMO}{Sequential Minimal Optimisation}
\acrodef{SOA}{Service-Oriented Architecture}
\acrodef{SOAP}{Simple Object Access Protocol}
\acrodef{SOC}{Self-Organised Criticality}
\acrodef{SOLUTA}{SOLUTA.NET}
\acrodef{SOM}{Self-Organising Map}
\acrodef{SSL}{Semantic Service Language}
\acrodef{STU}{Salzburg Technical University}
\acrodef{SUN}{Sun Microsystems}
\acrodef{SVM}{Support Vector Machine}
\acrodef{TM}{Turing Machine}
\acrodef{UBHAM}{University of Birmingham}
\acrodef{UDDI}{Universal Description Discovery and Integration}
\acrodef{UML}{Unified Modelling Language}
\acrodef{URI}{Uniform Resource Identifier}
\acrodef{UTM}{Universal Turing Machine}
\acrodef{VLP}{variable length population}
\acrodef{VLS}{variable length sequences}
\acrodef{vls}{variable length sequence}
\acrodef{WP}{Work-Package}
\acrodef{WSDL}{Web Services Definition Language}
\acrodef{XMI}{XML Metadata Interchange}
\acrodef{XML}{eXtensible Markup Language}
\acrodef{MD5}{Message-Digest algorithm 5}
\acrodef{GA}{genetic algorithm}
\acrodef{GP}{genetic programming}
\acrodef{MASON}{Multi-Agent Simulator Of Neighbourhoods}
\acrodef{Repast}{Recursive Porous Agent Simulation Toolkit}
\acrodef{JCLEC}{Java Computing Library for Evolutionary Computing}
\acrodef{OWL-S}{Web Ontology Language - Service}
\acrodef{EGT}{Evolutionary Game Theory}
\acrodef{RBF}{Radial Basis Functions}
\acrodef{SWS}{Semantic Web Services}

%% file: captions.tex
\acrodef{archComTop}{many strongly connected clusters (communities), called {sub-networks} (quasi-complete graphs), with a few connections between these clusters (communities) \cite{swn1}. Graphs with this topology have a very high clustering coefficient and small characteristic path lengths \cite{swn1},}
\acrodef{similarCap}{requests are evaluated on separate {islands} (populations), with their evolution accelerated by the sharing of solutions between the evolving populations (islands), because they are working to solve similar requests (problems).}
\acrodef{picCapUser}{will formulate queries to the Digital Ecosystem by creating a request as a {semantic description}, like those being used and developed in \aclp{SOA} \cite{SOAsemantic}, specifying an application they desire and submitting it to their habitat.}
\acrodef{picUserReq}{A population is then instantiated in the user's habitat in response to the user's request, seeded from the agents available at their habitat}
\acrodef{nlccap}{The agent life-cycle will change to support the {targeted migration}, as shown}
\acrodef{nlc2cap}{by the blue circle. Specifically, there will be more opportunities for agent migration, but more importantly these opportunities will be for {targeted migration}, which will help to optimise the set of agents found at the habitats}
\acrodef{svmCap}{A \acf{RBF} is recommended for text categorisation \cite{joachims1997tcs}, with the most common form of the \ac{RBF} being Gaussian \cite{gunn1998svm}.}
\acrodef{as3}{with an abstract representation consisting of a set of}
\acrodef{agentSemantic2}{tuple representing an {attribute} of the {semantic description}, one integer for the {attribute identifier} and one for the {attribute value}, with both ranging between one and a hundred.}
\acrodef{as4}{Each simulated agent had a semantic description}
\acrodef{semanticRequest}{A simulated user request consisted of an abstract {semantic description}, as a list of sets of numeric tuples to represent the properties of a desired business application}
\acrodef{bmlcap1}{the {numerical semantic descriptions}, of the simulated services (agents) and user requests, in a {human readable form}.}
\acrodef{capbml2}{The {semantic filter} translates {numerical semantic descriptions} for one community within the user base, showing it in the context of the travel industry}
\acrodef{capbml3}{The simulation still operated on the numerical representation for operational efficiency, but the {semantic filter} essentially assigns meaning to the numbers.}
\acrodef{tmcf}{The Digital Ecosystem alone averaged a 68.0\%}
\acrodef{tmc2f}{while the Digital Ecosystem with the {migration control} showed a significant degradation to 49.6\%}
\acrodef{tmc3f}{and the Digital Ecosystem with the {pattern recognition control} showed only a small increase to 70.5\%}
\acrodef{boohoo}{The Digital Ecosystem alone performed as expected, adapting and improving over time to reach a mature state}
\acrodef{nnbtmCap}{The Digital Ecosystem alone averaged a 68.0\% (3 s.f.) response rate with a standard deviation of 2.61 (2 d.p.), while the Digital Ecosystem with the \ac{NN}-based {targeted migration} showed a significant improvement to a 92.1\% (3 s.f.) response rate with a standard deviation of 2.22 (2 d.p.).}
\acrodef{tmsf}{The Digital Ecosystem with the \ac{SVM}-based {targeted migration} averaged a 92.8\% (3 s.f.) response rate}
\acrodef{tms2f}{slightly better than the \ac{NN}-based {targeted migration} at 92.1\% (3 s.f.)}
\acrodef{tms3f}{and so significantly better than the Digital Ecosystem alone at 68.0\% (3 s.f.)}
\acrodef{gSVM2cap}{Digital Ecosystem with the \ac{SVM}-based {targeted migration}, the Digital Ecosystem with the \ac{NN}-based {targeted migration}, and the Digital Ecosystem alone.}
\acrodef{gNN2cap}{The Digital Ecosystem alone performed as expected, adapting and improving over time to reach a mature state through the process of {ecological succession} \cite{begon96}}
\acrodef{gNN3cap}{Digital Ecosystem with the {targeted migration}, \ac{NN} or \ac{SVM}-based, showed a significant improvement}
\acrodef{histogram2Cap}{the greater effectiveness of the SVM-based {targeted migration}, compared to the NN-based {targeted migration}}
\acrodef{histogramCap}{frequency of poor matches ($<$50\%) every one hundred time steps, for the Digital Ecosystem with the SVM-based {targeted migration}, compared to the Digital Ecosystem with the NN-based {targeted migration}, and the Digital Ecosystem alone}